\documentclass[11pt,a4paper]{article}
\usepackage[hyperref]{naaclhlt2019}
\usepackage{times}
\usepackage{latexsym}

\usepackage{url}
\usepackage{latexsym}
\usepackage{url}
\usepackage{amsmath}
\usepackage{color}
\usepackage{makecell}
\usepackage{hyperref}
\usepackage{rotating}
\usepackage{multirow}
\usepackage{pbox}
\usepackage{caption}
\usepackage{tabularx}
\usepackage{float}
\usepackage{subfig}
\usepackage{caption}
\usepackage{tikz}
\usepackage{pgf-pie}
\title{Analyzing and Interpreting Convolutional Neural Networks in NLP}

  \aclfinalcopy

\author{Mahnaz Koupaee \\
  University of California, Santa Barbara \\
  {\tt koupaee@cs.ucsb.edu} \\\And
  William Yang Wang \\
 University of California, Santa Barbara \\
  {\tt william@cs.ucsb.edu} \\}
\date{}

\begin{document}
\maketitle
\begin{abstract}
Convolutional neural networks have been successfully applied to various NLP tasks. However, it is not obvious whether they model different linguistic patterns such as negation, intensification, and clause compositionality to help the decision-making process. 
In this paper, we apply visualization techniques to observe how the model can capture different linguistic features and how these features can affect the performance of the model. Later on, we try to identify the model errors and their sources. We believe that interpreting CNNs is the first step to understand the underlying semantic features which can raise awareness to further improve the performance and explainability of CNN models.
\end{abstract}

\section{Introduction}
Convolutional neural networks (CNNs) have been applied to multiple NLP tasks especially text classification, sentiment analysis \cite{johnson2014effective,kim2014convolutional} and more recently sequence-to-sequence language learning \cite{gehring2017convolutional,gehring2017convolutional2} and are able to match or outperform the state of the art systems. 
However, the underlying behavior of the CNNs is not easily interpretable. 

Neural networks including CNNs take the word embeddings as the input and operate on these real-valued vectors, pass them through some layers with hidden units and finally generate the output. 
Although the final performance of the model can be satisfying for the task under study, it is not well clear if the model can filter out the unnecessary information and capture the semantic features such as the effect of negation, different intensity levels implied by different terms and polarity composition of multiple clauses of the input text. 
Tuning the parameters of the model can lead to improved results however it is not obvious what new language features are captured by the model and how the changes can improve the overall performance of the model. 
Once the features captured by the CNN model are identified, the error sources can be recognized.
Knowing what causes the model to fail can reveal the limitations of neural models and also raise awareness on how to further improve the performance of future designs.


Recently, visualization as a means of exploring the behavior of neural models has been used for NLP tasks
followed by similar works in computer vision \cite{weinzaepfel2011reconstructing,vondrick2013hoggles,mahendran2015understanding,zeiler2014visualizing} and primarily focused on observing the semantic properties of the model.
Recurrent Neural Networks (RNNs) have been the most favorite architectures for analysis started by \citet{li2015visualizing} and \citet{karpathy2015visualizing} for NLP tasks, focusing on the learned semantic patterns. 
These methods only consider the association of the input tokens with the final output based on word embeddings and they do not reflect the structural information and model-specific parameters in the outcome.  

Moreover, CNNs as common architectures for various NLP tasks, have not been thoroughly explored yet. The existing works on CNNs visualization also apply similar strategies of sequence-to-sequence models, mainly the first-derivative saliency scores, to determine tokens contribution \cite{aubakirova2016interpreting,karlekar2018detecting}. 
CNNs have their unique properties: convolutional values calculated in each convolution layer, different number of feature maps, different kernel sizes and different stride sizes and none of them are studied for their effects on semantics in previous works.


In this paper, we make use of visualization techniques to interpret the CNN ability to achieve semantic properties.  
We start by specifying the most contributing terms to the final decision of the model. Next, we averaged the contributions of different parts of speech tags and identified the most contributing groups of words.
Later, different common semantic features such as negations, contrastive terms, compositionality, etc. are introduced and the model behavior is visualized in different cases. 
Finally, by analyzing failure cases, different error categories are identified and their distributions are computed.
The contribution of this work is three-fold: 
\begin{itemize}
\item We introduce new visualization techniques to interpret the underlying semantic behavior of CNNs with regard to their unique properties.
\item Using visualization techniques, we show what semantic features CNNs capture and how these features help them accomplish their goals for different NLP tasks.
 \item We analyze model failures to find the underlying error causes for such mistakes and the model limitations, hopefully guiding us for better future design.
\end{itemize}

\begin{figure*}
\centering
\begin{minipage}{.32\textwidth}
  \includegraphics[width=0.95\linewidth]{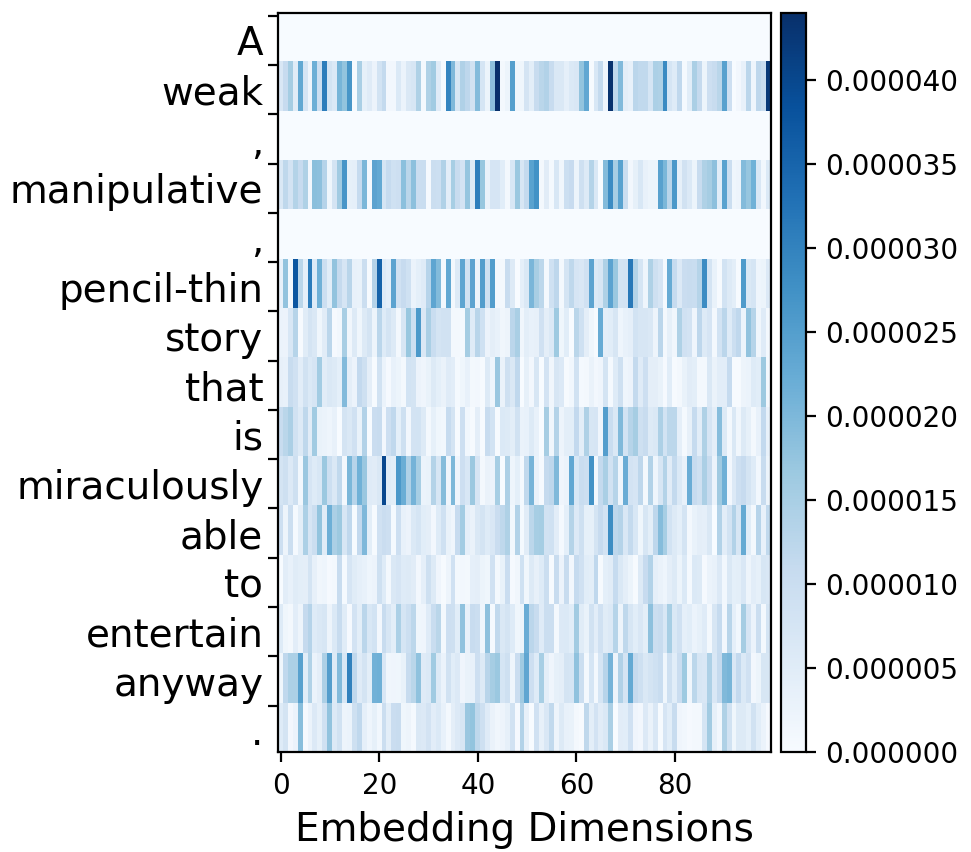}
  \label{fig:a}
  \caption*{a: saliency score visualization}
\end{minipage}
\begin{minipage}{.32\textwidth}
  \includegraphics[width=0.95\linewidth]{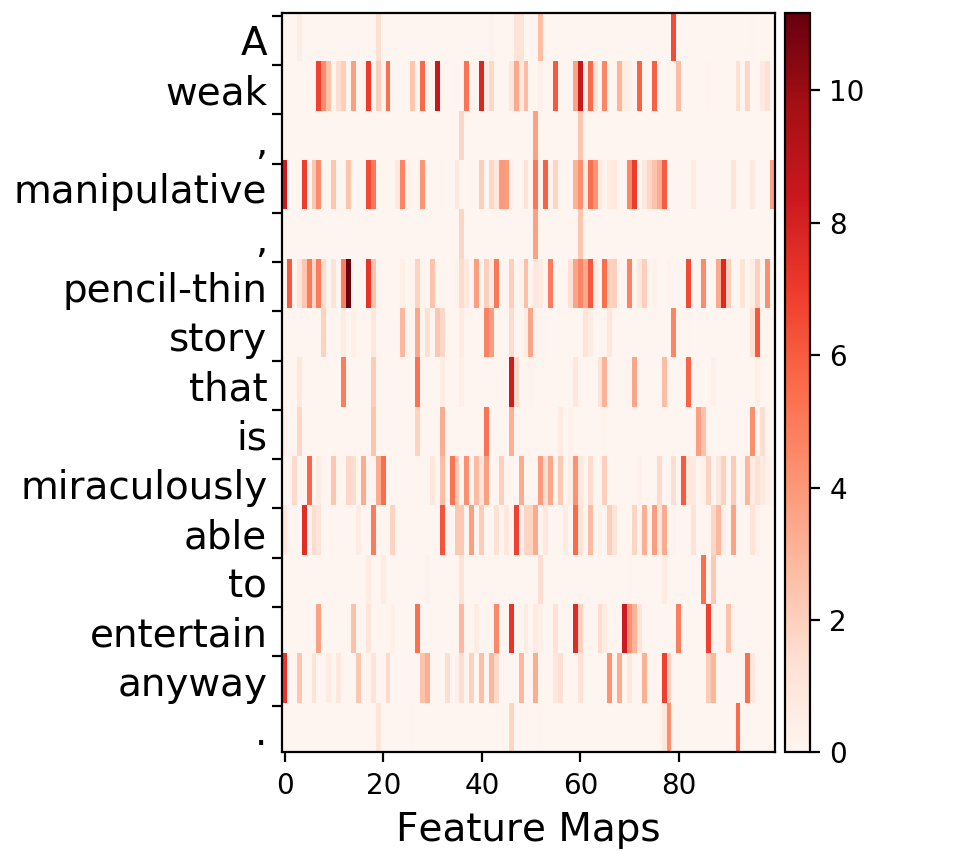}
  \label{fig:b}
  \caption*{b: convolution values visualization}
\end{minipage}
\begin{minipage}{.32\textwidth}
  \includegraphics[width=0.95\linewidth]{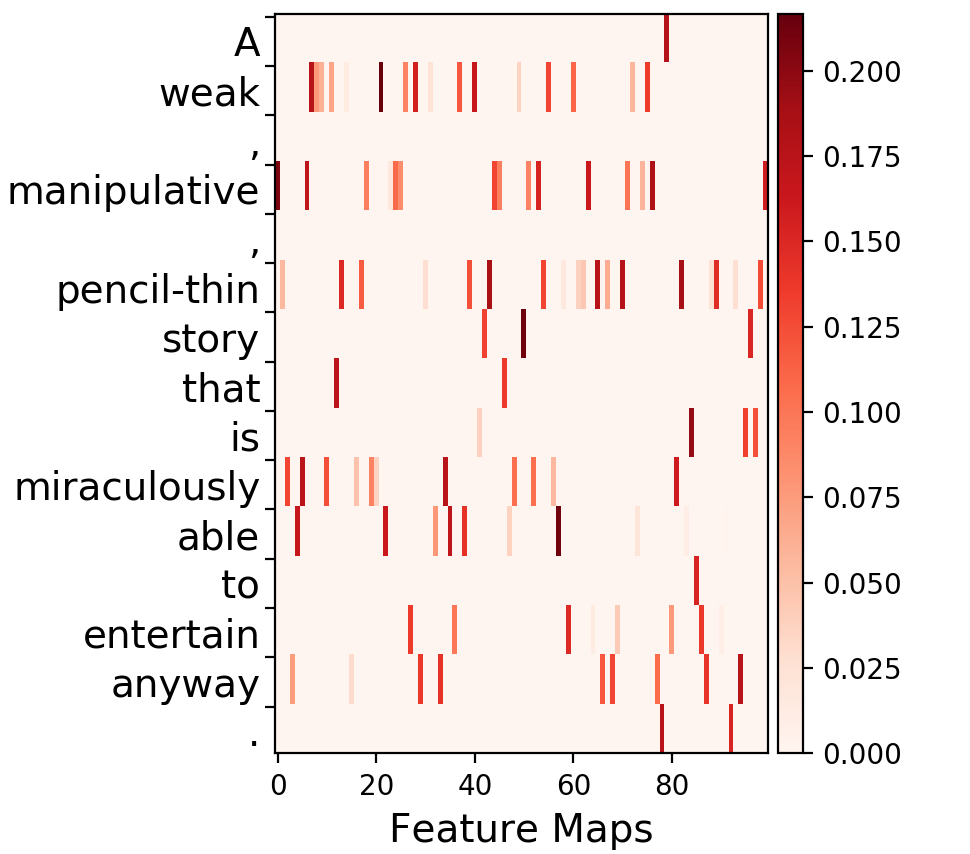}
  \label{fig:c}
  \caption*{c: linear weights visualization}
\end{minipage}
\label{fig-1}
\caption{Visualization based on different strategies. \textit{a} is a heat map representing the saliency score computed as the derivative of the final prediction score with respect to the word embeddings. \textit{b} is the proposed visualization based on the convolution values of different feature maps. \textit{c} shows the tokens dominance in the final linear layer.}
\end{figure*}

\begin{figure}
\begin{minipage}{.23\textwidth}
  \centering
  \includegraphics[width=0.81\linewidth]{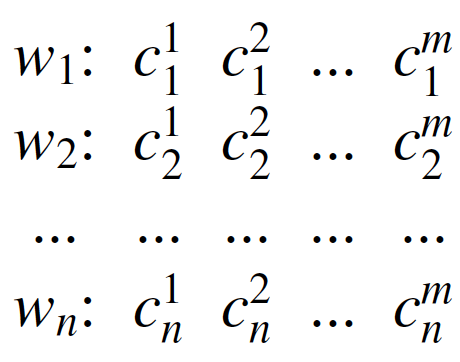}
  \caption*{{\scriptsize Single words representation}}
\end{minipage}
\begin{minipage}{.23\textwidth}
  \centering
  \includegraphics[width=0.81\linewidth]{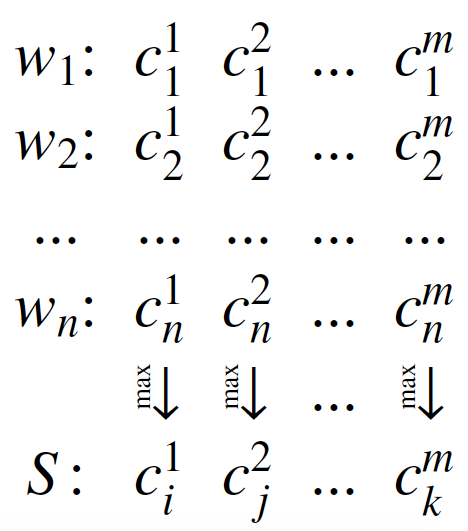}
  \caption*{{\scriptsize Phrase representation}}
\end{minipage}
\label{fig2}
\caption{Word and phrase visualization. Each word is represented by its convolution values in different feature maps. To represent a phrase or a sentence with a single vector, the maximum convolution values of feature maps are used.}
\end{figure}


\section{Related Work}
Neural architectures visualization techniques were initially applied for computer vision applications. Projecting outputs from different neural levels back to the initial input images is one of the early techniques used by \citet{weinzaepfel2011reconstructing} to visualize how different layers perform. \citet{vondrick2013hoggles} later used the same technique to visualize feature spaces used by object detectors. \citet{mahendran2015understanding} also compute an approximated inverse of the image to analyze deep CNNs. The deconvolutional networks presented by \citet{zeiler2014visualizing} explores the layers of a CNN by applying strategies to spot active regions contributing the most to the final classification decision.

Visualization is also shown to be useful for understanding neural networks for linguistic tasks. \citet{karpathy2015visualizing} use visualization to understand the hidden states in language models and demonstrates how cells can model language properties.
\citet{li2015visualizing} make use of gradient-based saliency scores to find important words of the input and later use this representation to analyze the linguistic patterns learned by RNN models. 
\citet{kadar2017representation} show that RNNs specifically learn lexical categories and grammatical functions that carry semantic information, partially by modifying the inputs fed to the model. 
These works represent how each input token is associated with the final prediction of the model however, the visualization does not give insight into the effect of different model-specific parameters.

A few works on CNN visualization also applied the saliency scores. \citet{aubakirova2016interpreting} use the politeness prediction task to visualize what CNNs are learning. \citet{karlekar2018detecting} also use the saliency heat maps to discover linguistic patterns of Alzheimer's disease patients. However, these works also ignore different parameters of CNNs for semantic visualization.

Designing interactive visualization tools to observe what is happening in neural networks has been of much interest recently.
CNNVis \cite{liu2017towards} models a deep neural network as a directed acyclic graph and applies several techniques to scale to large networks.  ActiVis \cite{kahng2018cti} is another interactive visualization tool that uses computational graphs to present the structure of deep neural networks.
LSTMVis \cite{strobelt2018lstmvis} and the SEQ2SEQ-VIS \cite{strobelt2018seq2seq} are two interactive visualization tools that can be used for sequence-to-sequence models.
These tools allow users to freely inspect different aspects of a neural network and learn about the role of each neuron and how they interact with each other. They are mainly aimed at helping users observe the effect of different choices for different parameters on the performance of the model.

We propose a visualization approach to illustrate how CNNs capture linguistic properties, not only by showing the contribution of the input tokens to the output, but also by showing this association with regard to different model parameters.

\section{CNN Visualization}
Visualization is the first step to help us figure out what semantic features neural models can learn for an NLP task. Behavior analysis of the model and gaining linguistic insights depend on this step.
\citet{li2015visualizing} made use of ``saliency score'' approximated by first derivatives with regard to the word embeddings to measure how much each input unit contributes to the final decision. If $S_c$ is the function to predict the class $c$ and $e$ represents the word embeddings, then the saliency score of each token will be calculated as follows:
\[
S(e) = |\frac{\partial S_c}{\partial e}|
\]
The final calculated score represents the sensitivity of the output prediction score with respect to the changes in different embedding dimensions. 
Using the same strategy for a CNN model, the resulting heat map of an example sentence is illustrated in \hyperref[fig:a]{Figure 1.a}. Each horizontal bar represents the calculated saliency scores for different embedding dimensions which shows the sensitivity of the final prediction to the changes in specific embedding dimensions of a word. 

\begin{figure*}
\centering
\begin{minipage}{.97\textwidth}
  \centering
  \includegraphics[width=0.9\linewidth]{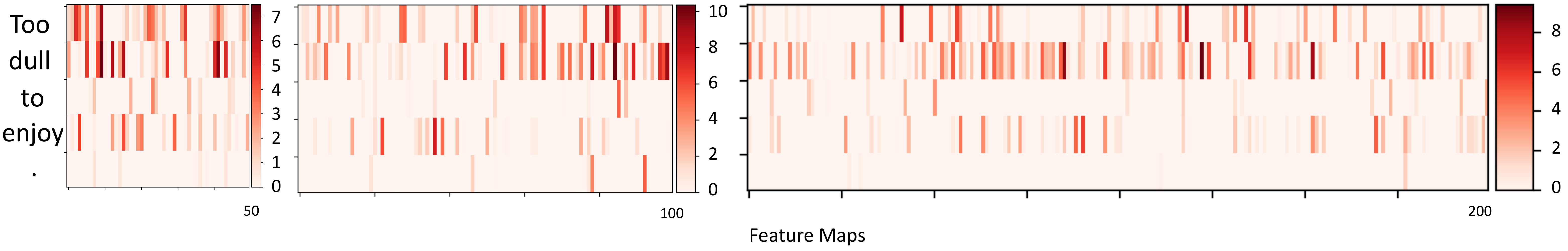}
  \label{fig:aa}
  \caption*{a: Visualization for different feature maps sizes}
\end{minipage}
\begin{minipage}{.8\textwidth}
  \centering
  \includegraphics[width=0.9\linewidth]{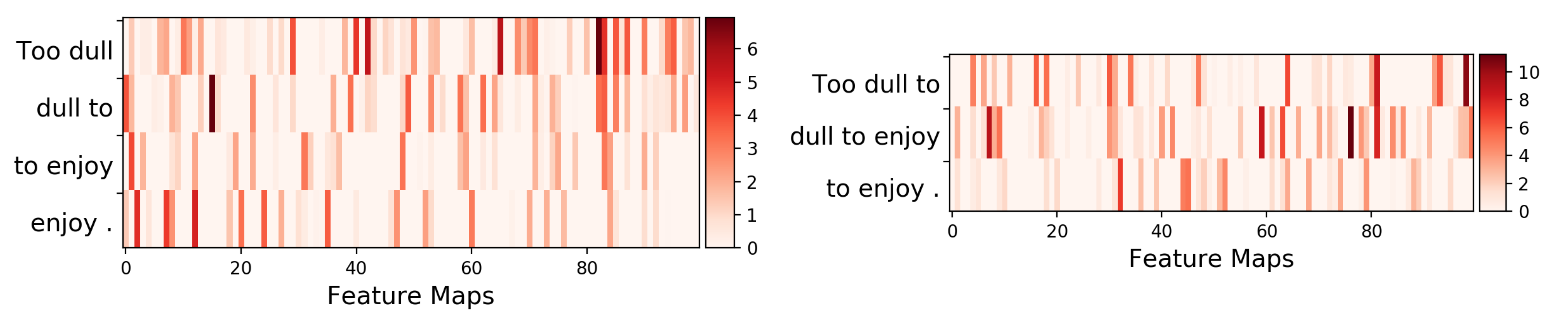}
  \label{fig:bb}
  \caption*{b: Visualization for kernel sizes of 2 and 3}
\end{minipage}
\label{fig-comb}
\caption{Multiple visualizations of a sentence with different parameters.}
\end{figure*}

A CNN model receives a sentence $S$ consisting of $n$ words, $w_1$, $w_2$, $\cdots$, $w_n$, where each word is a $d$ dimensional vector. A number of filters parameterized with $W_m$ is applied on the window of $t$ words and a convolution feature is produced by $c_i = f(W_m.w_{i:i+t-1} + b)$ with $f$ as a non-linear function. A feature map $c$ for a filter is then $[c_1,c_2,....,c_{n-t+1}]$ for all possible windows of words. The convolution layer is then followed by a max-pooling layer.

Calculating the first-derivative saliency scores with respect to the word embeddings can represent tokens contributions to the final decision of the model. However, not much information can be obtained about the structure of the CNN and its parameters, since the saliency scores are calculated over all the existing feature maps and with disregard to the feature maps numbers and kernel sizes.
While the saliency score can be computed for each feature map or with respect to the other parameters of the model and at different layers of the CNN, it requires much more effort and depending on the number of feature maps and convolution layers, there could be many heat maps to analyze.

To visualize the learned patterns by a CNN model, we applied a different yet simpler approach. Instead of calculating the saliency scores, we make use of the feature maps generated by filters applying convolution on word embeddings. Words are represented by $m$ dimensional heat map vectors ($c^1_i,c^2_i,...,c^m_i$) where $m$ is the number of filters and each $c^j_i$ represents the value of a convolution feature for token $i$ in feature map $j$. A phrase consisting of multiple tokens is also shown by a vector consisting of maximum convolution values for each feature map.
\hyperref[fig2]{Figure 2} shows how we represent single words and phrases using the convolution values.

Applying the proposed method to the same example, the generated heat map is shown in \hyperref[fig:b]{Figure 1.b}. Each row represents the computed convolution values of a specific token using different feature maps. \hyperref[fig:a]{Figure 1.a} and \hyperref[fig:b]{Figure 1.b} show similar results as in both heat maps the most contributing tokens are similar. Later, we propose a way to show that our approach and the one based on saliency score result in similar output.

We use the computed values at the convolution layer to illustrate the terms contribution, however one might claim that the final linear layer is likely to assign lower values to the tokens considered as the most contributing ones, therefore reducing their effects. As shown in \hyperref[fig:c]{Figure 1.c}, the assigned weights in the linear layer are also compliant with the convolution weights.

Finally, \hyperref[fig-comb]{Figure 3} show how visualization based on convolution values can reflect different CNN parameters. The examples represent the heat maps of a sentence with respect to different feature maps and kernel sizes. The same strategy can be applied for different stride sizes and different layers, depending on what we want to analyze.

\section{Datasets and the CNN Model}
The approach in this paper can be generalized to be used for any NLP task and dataset, however to better illustrate the model analysis, examples are used throughout the paper. Therefore, we selected two datasets and three different classification tasks to show how CNNs can capture meaning. 
Stanford Sentiment Treebank (STT) is a benchmark dataset used for sentiment analysis task. The dataset contains phrases
of 11,855 sentences and there exists both fine-grained (very positive, positive, neutral, negative and very negative) and coarse-grained (positive, negative and neutral) ground-truth labels for every phrase. We made use of the scripts available by \citet{socher2013recursive} to divide the dataset into train, validation and test sets. To perform the coarse-grained classification, pairs with neutral labels are removed. 
The Subjectivity Dataset is a classification dataset \cite{pang2004sentimental} containing 9,000 sentences where the task is to classify sentences as being subjective or objective. We made use of this dataset along with the STT to visualize multiple features CNN can learn and capture for different tasks. The data statistics are shown in \hyperref[data]{Table 1}. 

Our CNN model consists of one convolutional layer with kernel and stride size of $1$ and $100$ different filters followed by a max-pooling layer. 
Finally to classify the sentences, the outputs of theses layers are concatenated and fed to a linear layer.
The word embeddings are trained from scratch and the validation set is used for parameter tuning and early stopping. The performance of this model on different tasks are shown in \hyperref[data]{Table 1}.

\begin{table}
\scriptsize
\centering
\begin{tabular}{lccc}
&\makecell{Coarse-grained \\STT} & \makecell{Fine-grained \\ STT} & \makecell{Subjectivity\\ Dataset}\\
\hline
Training Set Size&117,220&236,076&8,000\\
Validation Set Size&825&1,044&500\\
Test set size&1,749&2,125&500\\
\hline
Accuracy on Validation&91.15\%&52.78\%&88.80\%\\
Accuracy on Test&93.20\%&51.06\%&89.00\%\\
\hline
\end{tabular}
\caption{The statistics of the datasets and the performance of the model on different tasks.}
\label{data}
\end{table}

\begin{figure*}
\centering
\begin{minipage}{.32\textwidth}
  \includegraphics[width=0.95\linewidth]{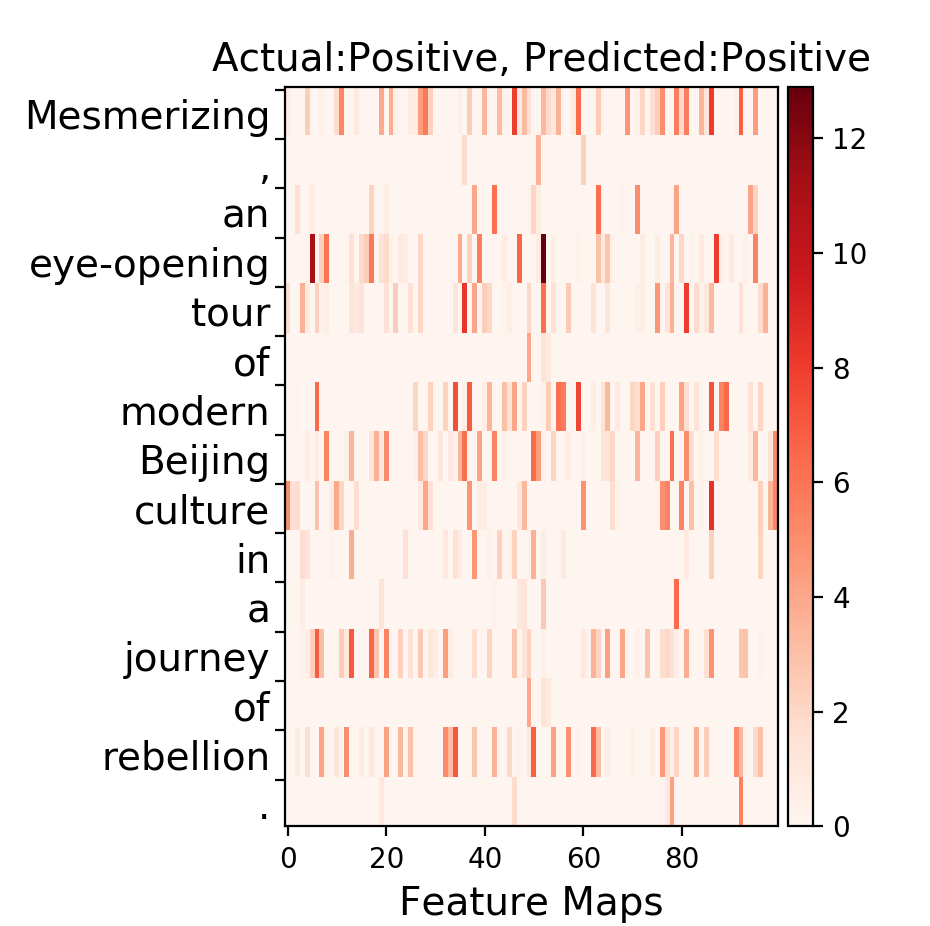}
  \label{fig:test1}
\end{minipage}
\begin{minipage}{.32\textwidth}
  \includegraphics[width=0.95\linewidth]{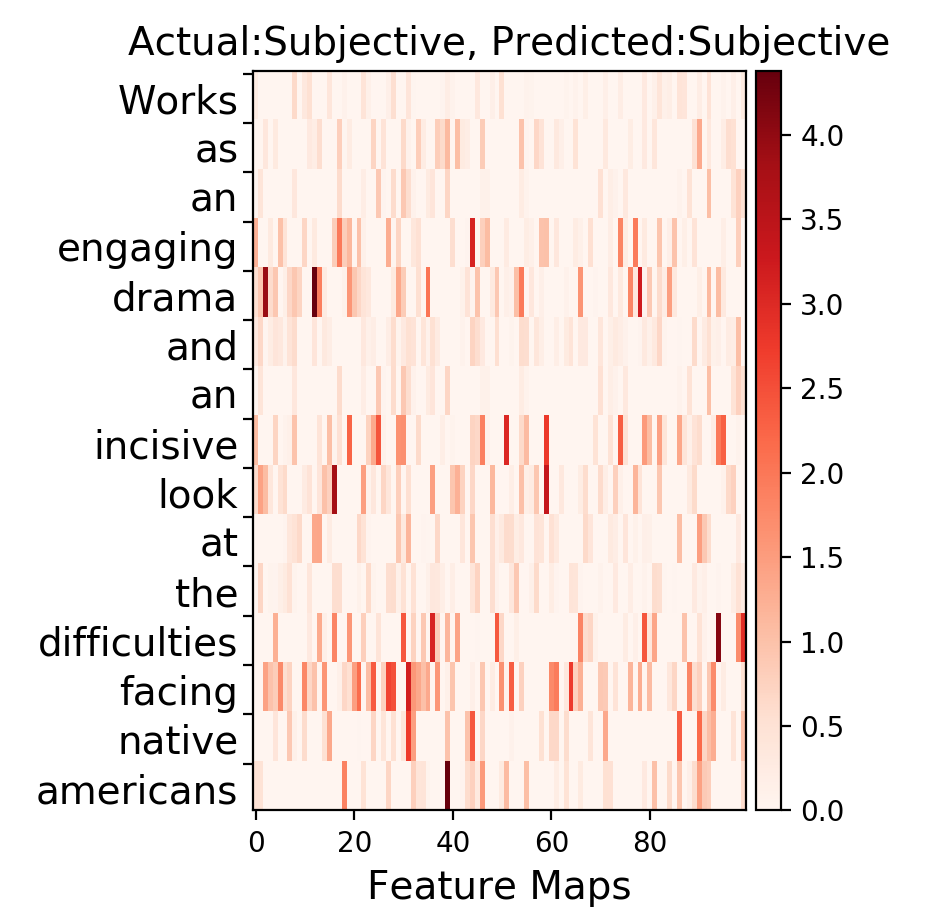}
\end{minipage}
\begin{minipage}{.32\textwidth}
  \includegraphics[width=0.95\linewidth]{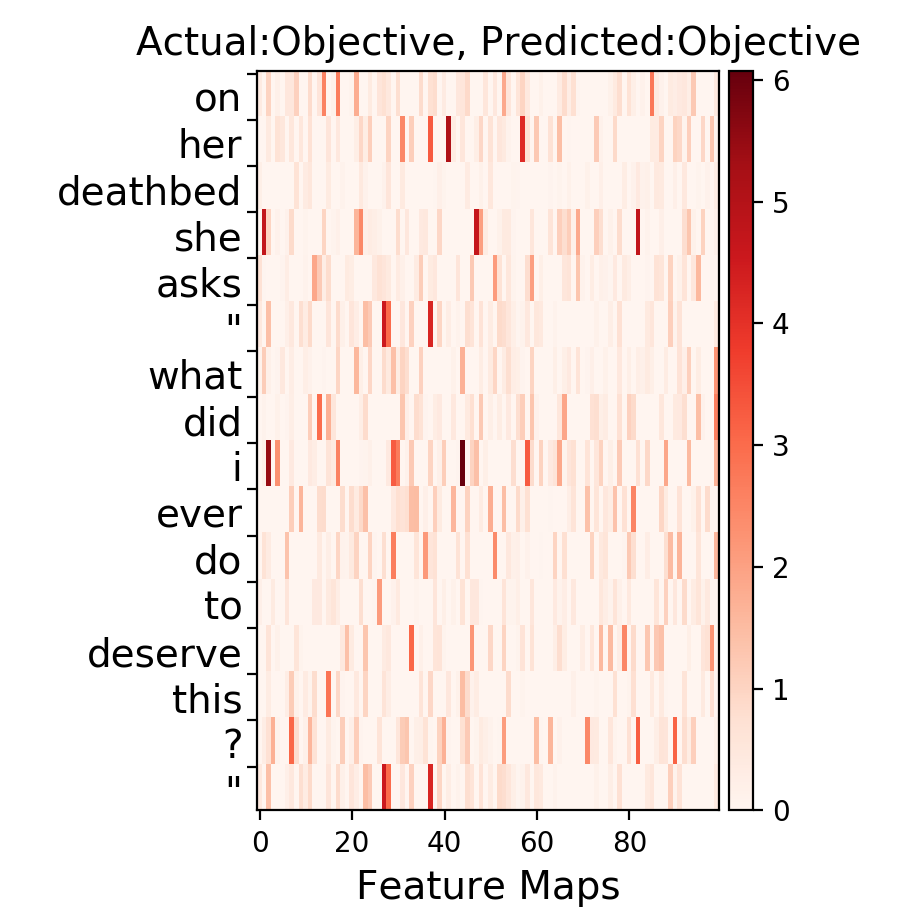}
\end{minipage}
\label{fig1}
\caption{Terms contribution to predicting the final output. The first figure is from the STT dataset and the other two are from Subjectivity (Subj) dataset.
Determiners and punctuation contribute the least while adjectives and adverbs are the most dominant features for the first task and also deciding on the subjectivity of a sentence for the second task. However,  pronouns (I, her) and punctuation (``, ?) are among the most contributing features for identifying objective sentences.}
\end{figure*}

\section{CNN Analysis}
Visualization is aimed at helping realize what semantic features CNNs can capture and how they achieve them. We selected NLP common linguistic features introduced in \cite{li2015visualizing}. 
Each entry in the dataset consists of words with different parts of speech. We start by showing their contributions and finding the most contributing groups of words for each task. Later, contrast, negation, intensity and compositionality are analyzed.

Once again, it is worth mentioning that throughout the following sections, different examples from our datasets have been used to better elucidate how the model captures the linguistic features with the visualization technique. However, the use of specific examples in different sections does not limit the generalizability of the technique.

\subsection{Dominant Groups of Tokens}
\begin{table*}
\scriptsize
\centering
\begin{tabular}{lp{4cm}|p{4cm}|p{4cm}}
&\makecell{Coarse-grained \\sentiment analysis} & \makecell{Fine-grained \\sentiment analysis} & \makecell{Subjectivity/Objectivity \\classification}\\
\hline
Convolution value based&ADJ/ ADV/ NOUN/ VERB/ NUM/ X/ PRON/ ADP/ CONJ/ PRT/ DET/ . 
&ADJ/ ADV/ NOUN/ VERB/ X /NUM /PRON/ ADP / ./ CONJ /DET/ PRT
&PRON/ ADV/ NOUN/ ADJ/ NUM/ VERB/ ADP/ X/ CONJ/ PRT/ ./ DET
\\
\hline
Saliency score based&ADJ/ ADV/ NOUN/ VERB/ NUM/ X/ PRON/ ADP/ CONJ/ ./ DET/ PRT/&
ADJ/ ADV/ NOUN/ VERB/ NUM/ X /PRON/ CONJ/ ADP / ./ DET/ PRT&
PRON/ ADV/ NOUN/ ADJ/ NUM/ VERB/ X/ ADP/ CONJ/ PRT/ ./ DET
\\
\hline
\multicolumn{4}{p{16cm}}{{Universal Part-of-Speech Tagset:\newline ADJ:adjective, ADP:adposition, ADV:adverb, CONJ:conjunction, DET:determiner/article, NOUN:noun, NUM:numeral, PRT:particle, PRON:pronoun, VERB:verb, .:punctuation marks, X:other}}\\
\hline
\end{tabular}
\caption{The dominant groups based on convolution value and saliency scores, ordered for different tasks. The most dominant groups for the first and second tasks are adjectives and adverbs, while the pronouns play the most important role for the subjectivity/objectivity task. The dominant sets are almost the same using either method, showing our method is able to determine tokens contributions similar to the saliency-based visualization.}
\label{dom}
\end{table*}

The first step to analyze the behavior of a CNN model is to see how much different tokens contribute to the output of the model and then identify the dominant groups for different tasks. 
To predict the model prediction for an entry, the input tokens are fed to the trained model. Using a fixed number of filters ($100$ in this case), the outputs of the convolution layer for all input tokens are calculated. 
The heatmaps of the input tokens are calculated as shown in \hyperref[fig2]{Figure 2} to show how important each word is for the final decision. \hyperref[fig1]{Figure 4} shows examples from each dataset. Each row represents the convolution features of a specific token for all feature maps. Tokens having the highest values in most of the feature maps have a stronger contribution for predicting the final output. 

Each word has a role in a sentence known as part of speech (noun, verb, adjective, etc.). Different groups of words can have different effects for a specific task. For instance, in the first example selected from the film review dataset shown in \hyperref[fig1]{Figure 4}, the model assigns higher importance to adjectives and adverbs, suggesting these groups of words have the highest impact on the decision made by the CNN model. 
To determine the dominant word groups for each task and dataset, we initially calculated the convolution values for all the entries in the test set and normalized the values for the tokens in each entry to be in range zero to 1. Next, using NLTK tagger, the part-of-speech of all words are determined. The universal tag set consisting of 12 different roles is used. Once the tokens' parts of speech are identified, the convolution values for each separate role is averaged over all entries of the test set. The parts of speech with highest average are the most dominant groups.  

For the review sentiment analysis task, the adjectives and the adverbs are the most effective ones while the pronouns are not even in the top 5 groups. The subjective/objective classification task requires deciding whether the sentence is describing the movie (subjective) or describing the plot (objective). To determine the subjectivity/objectivity of the sentences, pronouns have the highest convolution values along with adjectives, nouns and adverbs. The complete order of different groups for each task is presented in  
\hyperref[dom]{Table 2}.

Previously, the differences of the saliency-based visualization method and our CNN-specific visualization technique were discussed. It is claimed that the difference of the techniques is merely representational but their outcomes illustrating the terms contributions are similar. To show the symmetry between the results of two methods, we identified the dominant groups for each dataset and task as above both by computing the first derivative saliency scores and convolution values. The results of \hyperref[dom]{Table 2} show that the most dominant groups identified by both methods are similar.

\subsection{Contrast and Negation}
To distinguish the opposite polarities (positive vs negative) and sentences with contrasting meanings , it is necessary for a CNN model to capture contrasting words either in form of the opposites (interesting vs boring) or in the existence of negations (like vs don't like). 

\begin{figure}
\subfloat{%
  \includegraphics[clip,width=\columnwidth]{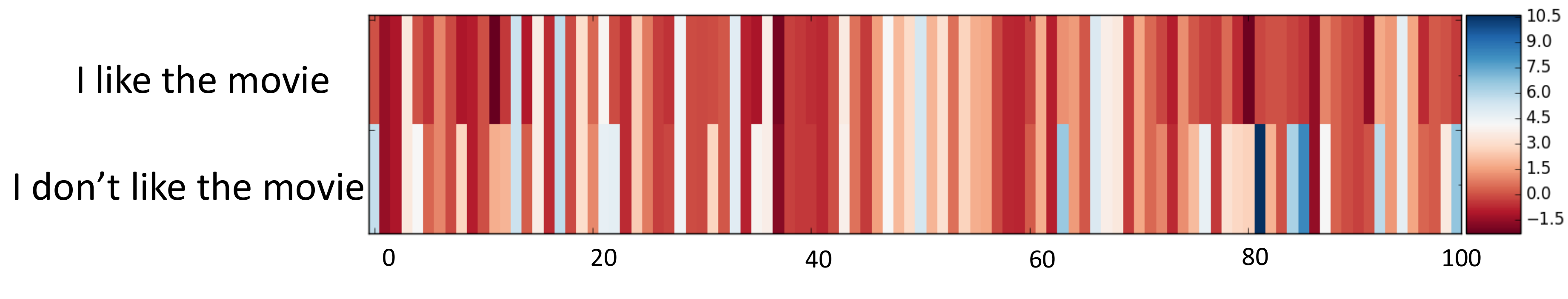}%
}
\vspace{-5mm}
\subfloat{%
  \includegraphics[clip,width=\columnwidth]{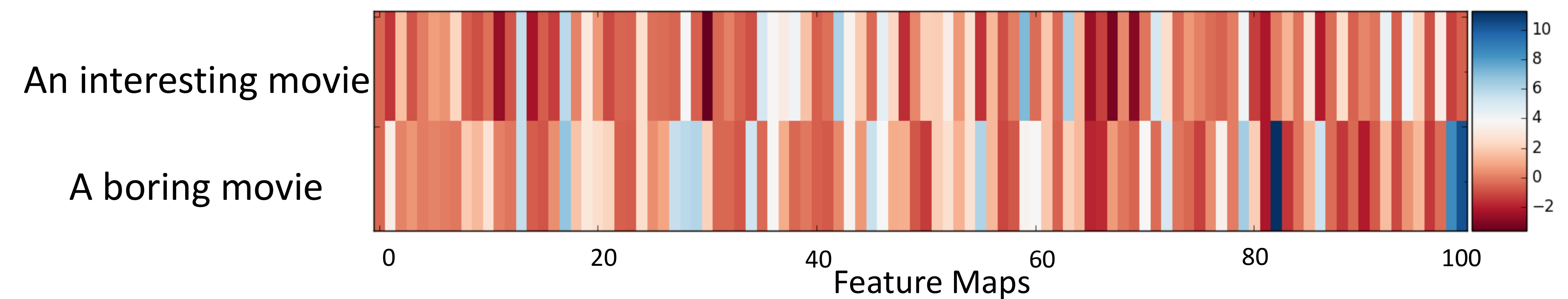}%
}
\label{cont}
\caption{Contrast and negation representation. The weights of the sentences with negative polarity are stronger compared to the positive ones.}
\end{figure}

To visualize these properties, for each phrase or sentence, the visualization vectors are calculated as depicted in \hyperref[fig2]{Figure 2}. 
\hyperref[cont]{Figure 5} illustrates two groups of contrasting examples. For the first group, the two sentences only differ in one token ("not") but with completely opposite meanings. To be successful in making the correct decision, the CNN model needs to distinguish the opposite polarities by capturing "not" in the first group and discriminating "interesting" and "boring" in the second group. As it can be seen, in each group, there are reversals in the weights for the sentences with opposite polarities. One thing to note is that for all diagrams in this paper we use the absolute values but for showing contrast we used the real values to illustrate the existing reversals.

\subsection{Intensity}
Another important pattern in language is the intensification which helps decide on the polarity level (for instance, \textit{positive}, \textit{very positive}) for the non-binary classification tasks. "Breathtaking" is much more stronger than "good". The intensification can be either represented by adverbs with different polarities like "really" or "amazingly" or can be inherent in a term such as using "terrible" instead of "bad". 

\begin{figure}
\subfloat{%
  \includegraphics[clip,width=\columnwidth]{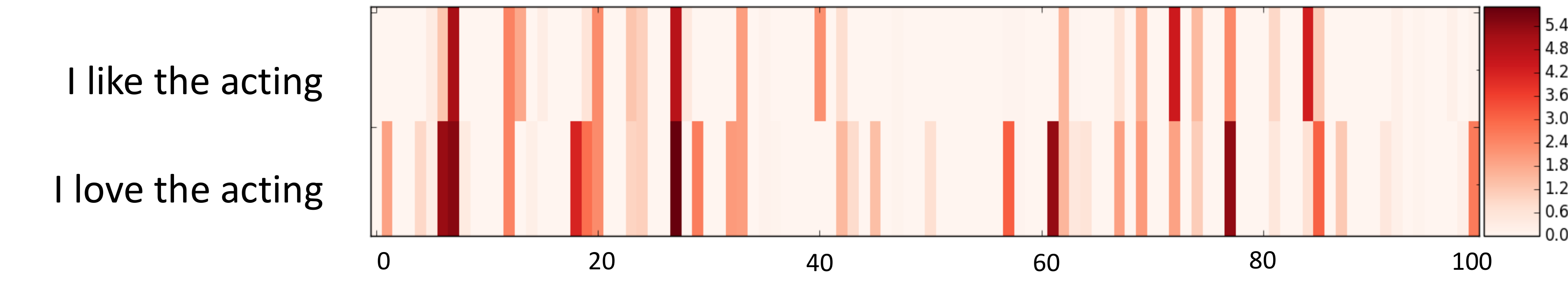}%
}
\vspace{-5mm}
\subfloat{%
  \includegraphics[clip,width=\columnwidth]{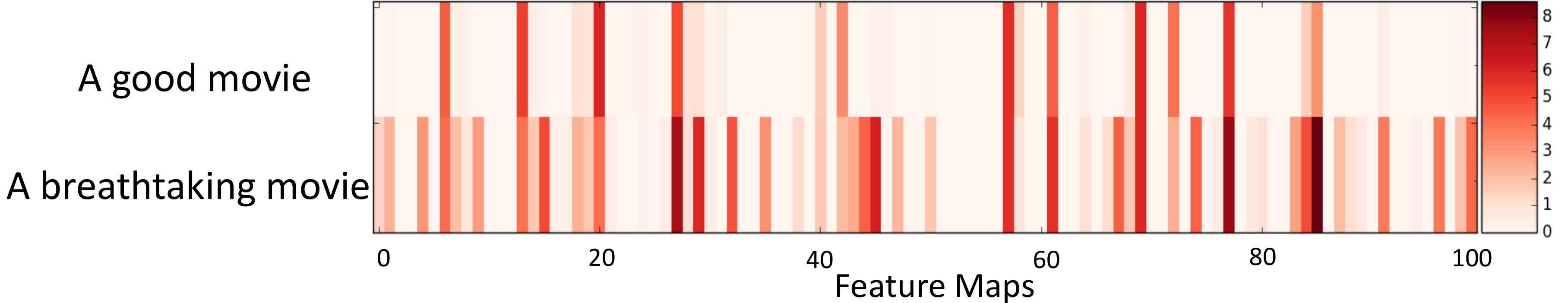}%
}
\label{intense}
\caption{Intensity representation. The weights of the sentences are stronger compared to the not intensified ones resulting in stronger polarities.}
\end{figure}

\hyperref[intense]{Figure 6} represents two sets of examples in which the sentence with an intensifying term has higher convolution values and thus higher final impact on the decision of the model. As stated earlier, the ability of a CNN model to capture intensification is most important in non-binary classification. For the examples in \hyperref[intense]{Figure 6}, due to the higher weight of "love" and "breathtaking", the CNN model predicts \textit{positive} for the first sentences and \textit{very positive} for the second ones.

\subsection{Compositionality}
The existence of different clauses and their compositions might make the prediction task of the model more challenging. Different conjunctions can be applied to connect two or more sentences with either compliant or contrasting meanings to form a sentence as a whole. The final class or sentiment of the sentence depends on its clauses.

A compound sentence can have \textit{compliant clauses} with the same polarity describing a similar idea. On the other hand, \textit{contrasting clauses} are composed of parts representing opposite polarities. The ability of a CNN model to capture such relations is a key to its success for making the right decision. The following sections describe these two forms of compositionality and it is shown through visualization, how the CNN model captures the underlying relation of different clauses of a sentence.  

\subsubsection{Compliant Clauses}
\begin{figure}
\centering
  \includegraphics[width=\linewidth]{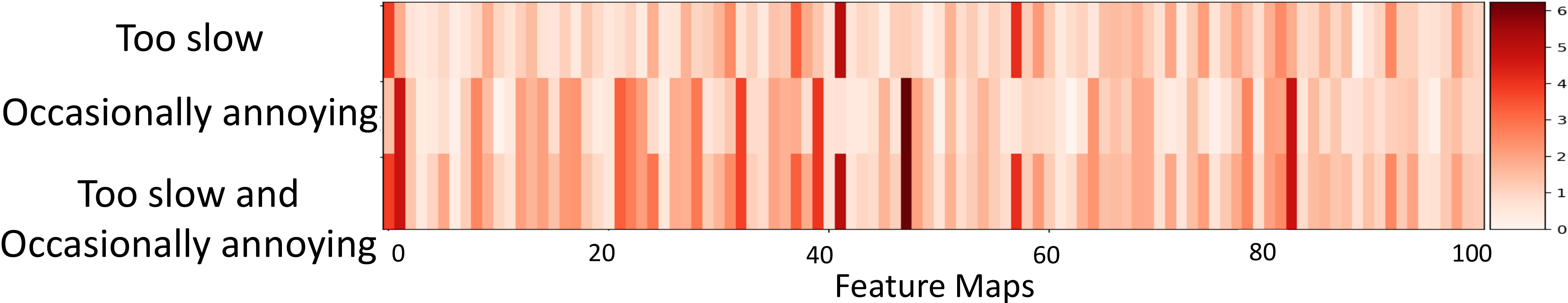}
\label{compliant}
\caption{Compliant clauses compositionality. The sentence combining two \textit{negative} clauses has higher weights and result in \textit{very negative} class.}
\end{figure}

Conjunctions such as ``and'', ``also'', etc. join two compliant clauses describing similar ideas much more strongly. ``Too slowly'' and ``occasionally annoying'' are two clauses the model predicts \textit{negative} as their labels. Interestingly, after combining these two clauses, the model successfully predicts \textit{very negative} for ``Too slowly \textbf{and} occasionally annoying''.  We used the visualization technique similar to the one used for contrast and intensity. \hyperref[compliant]{Figure 7} depicts that the sentence composing of two negative clauses have higher convolution values and therefore results in predicting \textit{very negative}.
This property is also observed for the subjectivity/objectivity task. ``Jamie has a to-do list for her life \textbf{and} also a very big secret she must keep from London .'' results in larger value for \textit{objective} class compared to its single objective clauses. 

\begin{figure}
\centering
  \includegraphics[width=\linewidth]{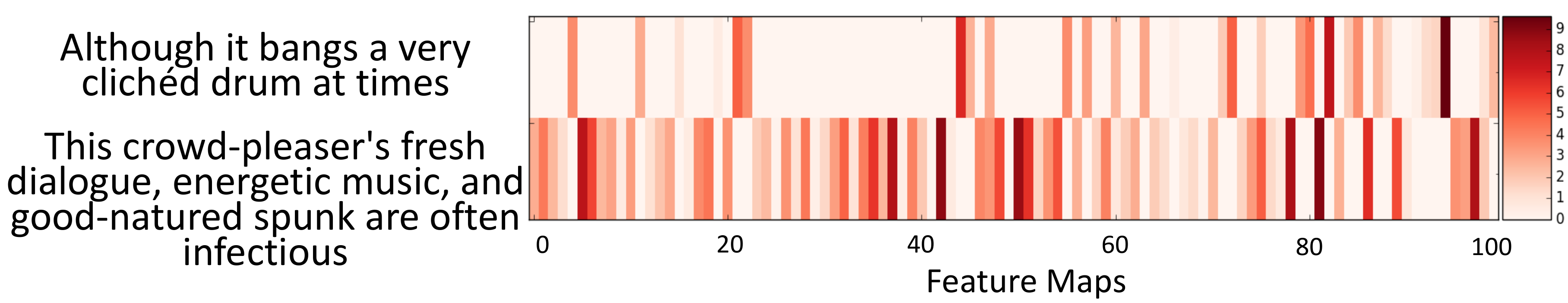}
\label{contrasting}
\caption{Contrasting clauses compositionality. Two clauses of a composite sentence. The \textit{positive} clause is more dominant having higher weights and therefore the model successfully predicts the \textit{positive} sentiment.}
\end{figure}

\subsubsection{Contrasting Clauses}
Contrasting clauses make things complicated as the model needs to weigh clauses and determines the most dominant one. 
The model is successful if it can determine the dominant clause. Knowing the classes of clauses, we want to visualize how the model can derive the final class. We first generate vectors for the clauses and the sentence combining them as depicted in \hyperref[fig2]{Figure 2}. Later, we calculate the difference of each clause vector with the sentence vector. The idea is that subtracting the values of the positive (or subjective) clause from the whole sentence will result in the effective negative (or objective) contributions of the other clause and subtracting the negative (or objective) clause values from the sentence will show the positive (or subjective) clause contribution. The one with higher contribution determines the class of the sentence. \hyperref[contrasting]{Figure 8} shows this process for ``\textbf{Although} it bangs a very clichéd drum at times, this crowd-pleaser's fresh dialogue, energetic music, and good-natured spunk are often infectious''.
This sentence consists of two clauses. The first one states the negative point about the movie, while the second clause specifies the positive points more emphatically.


\section{Error Analysis}
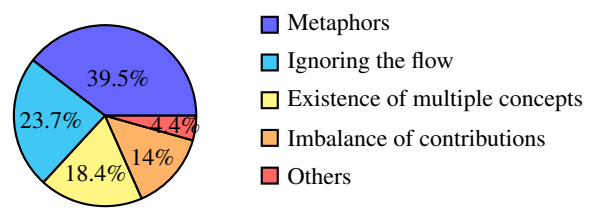
\begin{figure}
\centering
\small
\begin{tikzpicture}[scale=0.4]
\pie[text=legend]{
39.5/Metaphors, 23.7/Ignoring the flow, 18.4/Existence of multiple concepts, 14/Imbalance of contributions, 4.4/Others}
\end{tikzpicture}
\caption{Distribution of different error causes}
\label{errors}
\end{figure}

In previous sections, a number of common linguistic features are introduced and it is shown through visualization how the CNN model is able to capture such properties. 
Despite the ability of a CNN model to capture linguistic properties and using them to make the correct decision, there are cases where the model fails its task and generates the wrong output.  
Knowing what causes the model to fail is of much interest as it reveals the limitations of CNNs and helps improving future designs. 
For the purpose of this section, we selected the SST dataset and the binary classification task. Next, the cases in which the wrong output is generated by the CNN model are analyzed. Looking thoroughly at the cases where the model fails (6.8\% of total cases), We categorized the failures causes into five major groups.
The overall share of each error category is shown in \hyperref[errors]{Figure 9}. 
Please note that although the results of the error analysis conducted in this session is only for SST dataset, similar causes are the source of errors for other datasets and tasks.

Different failure causes are introduced through a few examples in the following sections. It is also worth mentioning that not all cases are 100\% distinct from each other and there may be the possibility of overlaps for some failure cases meaning that one sentence is misclassified as the result of multiple causes. 

\subsection{Existence of Metaphors} 
People have different writing styles. While some may be direct in saying what they mean, others adopt uncommon styles. It is common to use adjectives and adverbs while writing a piece of review to describe the quality of something. However, there are multiple cases where the true meaning behind the piece is not easily conveyed through the meaning of its single constituents.

``If Melville is creatively a great whale, this film is canned tuna.'' is comparing two films using symbols saying the film being reviewed is not good compared to another one. Figuring this out is easy for humans as they have seen and heard of this type of style a lot. Since the model is not much proposed to this type of sentences it is expected to fail in such cases. The model fails in this case and similar cases since it only considers the words individually. 

The existence of metaphors is the most common cause of failures as shown in fig \hyperref[errors]{Figure 9}. This is responsible for about 40\% of failures showing that further improvements of the model performance highly requires addressing this issue.

\subsection{Ignoring the Flow} 
This type of errors can have three different forms. The first failure case is due to model inability to capture the temporal flow of the sequence. Capturing the polarity change of ``a strong first quarter, slightly less so second quarter, and average second half.'' over time is not easy for the model, therefore causing it to predict the wrong output. 

The second case happens when the model fails to address the relationship of the elements of a piece. The term ``no'' in ``no number of fantastic sets, extras, costumes and spectacular locales can disguise the emptiness.'' makes the whole sentence negative. The CNN model predicts the wrong output as it cannot capture such relation.

The CNN model under study has a kernel size of $1$ and therefore considers each input token separately and therefore is not able to follow the flow in the sequence. Increasing the kernel size can improve the overall accuracy, but even with larger kernel sizes the model is prone to failure.

A sentence can be declarative, interrogative or exclamatory. Based on the type, the sentence class can be different. Some failures occur due to the inability of the model to distinguish sentence types. ``It isn't great.'' is classified as \textit{negative} correctly, however, the model can not correctly predict \textit{positive} for tag question ``Isn't it great?''


\subsection{Existence of Multiple Concepts} 
There are a good number of failures where the most contributing tokens are not describing the main concept. ``A resonant tale of racism, revenge and retribution.'' is classified as negative since the model assigns high weights to the adjectives describing what the story is about not the movie itself. 
Although one can easily realize ``resonant'' is the term we are looking for, the model is incapable of figuring this out. 
The CNN model does not care what each token is describing and therefore the effect of all the elements of a piece of text is counted towards the final prediction.


\subsection{Imbalance of Tokens Contributions}
The fourth group of errors happen due to the existing imbalances in the training set and therefore the way word embeddings are learned.
``A very average science fiction film.'' is predicted to be positive.
We have used the visualization technique and observed that the word ``science'' is mainly responsible for the prediction of \textit{positive} tag for this sentence. The reason is that the embeddings of this word is closer to the ones which make a sentence positive. 

There are also a very few number of errors where the causes do not fall into the existing categories. 
For some cases the ground truth label in the dataset does not seem to be correct, or it might be the case where the visualization is not able to capture the cause of failure. These cases are shown in the \hyperref[errors]{Figure 9} as others.

Finally, it is worth mentioning that despite different specified error categories, they all occur due to a common cause: The CNNs mostly work on the surface and are not able to dig inside the semantics and extract meanings. We believe that further improvements of CNNs for NLP tasks require deeper capturing of the underlying semantics.

\section{Conclusion}
In this paper, we used simple techniques to initially visualize and interpret properties CNNs can capture and later identify the cases they fail. This analysis, though limited, can be used as a guideline for researchers to make decisions with higher awareness considering the strengths and the weaknesses of the CNN models. 


\bibliography{m.bib}
\bibliographystyle{aaai.bst}
\end{document}